\documentclass[preprint]{elsarticle}

\usepackage{lineno}
\modulolinenumbers[5]
\usepackage{geometry}
\usepackage{graphicx}
\usepackage{amsmath}
\usepackage{multirow}
\usepackage{amssymb}
\usepackage{upgreek}
\newcommand{\myvec}[1]{\mathbf{#1}}

\makeatletter
\def\ps@pprintTitle{%
 \let\@oddhead\@empty
 \let\@evenhead\@empty
 \def\@oddfoot{}%
 \let\@evenfoot\@oddfoot}
\makeatother

\begin{document}

\begin{frontmatter}
\title{Using Deep Learning for price prediction by exploiting stationary limit order book features}

\author[auth]{Avraam Tsantekidis}\ead{avraamt@csd.auth.gr}
\author[auth]{Nikolaos Passalis}\ead{passalis@csd.auth.gr}
\author[auth]{Anastasios Tefas}\ead{tefas@aiia.csd.auth.gr}
\author[tutfi]{Juho Kanniainen}\ead{juho.kanniainen@tut.fi}

\author[tutml]{Moncef Gabbouj}\ead{moncef.gabbouj@tut.fi}

\author[aar]{Alexandros Iosifidis}\ead{alexandros.iosifidis@eng.au.dk}

\address[auth]{Department of Informatics, Aristotle University of Thessaloniki, Thessaloniki, Greece}
\address[tutfi]{Laboratory of Industrial and Information Management, Tampere University of Technology, Tampere, Finland}
\address[tutml]{Laboratory of Signal Processing, Tampere University of Technology, Tampere, Finland}
\address[aar]{Department of Engineering, Electrical and Computer Engineering, Aarhus University, Denmark}

\begin{abstract}
The recent surge in Deep Learning (DL) research of the past decade has successfully provided solutions to many difficult problems. The field of quantitative analysis has been slowly adapting the new methods to its problems, but due to problems such as the non-stationary nature of financial data, significant challenges must be overcome before DL is fully utilized. In this work a new method to construct stationary features, that allows DL models to be applied effectively, is proposed. These features are thoroughly tested on the task of predicting mid price movements of the Limit Order Book. Several DL models are evaluated, such as recurrent Long Short Term Memory (LSTM) networks and Convolutional Neural Networks (CNN). Finally a novel model that combines the ability of CNNs to extract useful features and the ability of LSTMs' to analyze time series, is proposed and evaluated. The combined model is able to outperform the individual LSTM and CNN models in the prediction horizons that are tested.
\end{abstract}


\end{frontmatter}


\section{Introduction}

In the last decade Machine Learning (ML) has been rapidly evolving due to the profound performance improvements that Deep Learning (DL) has ushered. Deep Learning has outperformed previous state-of-the-art methods in many fields of Machine Learning, such as Natural Language Processing (NLP)~\cite{deng2018feature}, image processing~\cite{larsson2018robust} and speech generation~\cite{van2016wavenet}. As the number of new methods incorporating Deep Learning in many scientific fields increase, the proposed solutions begin to span across other disciplines where Machine Learning was used in a limited capacity. One such example is the quantitative analysis of the stock markets and the usage of Machine Learning to predict price movements or the volatility of the future prices or the detection of anomalous events in the markets.

In the field of quantitative analysis, the mathematical modelling of the markets has been the de facto approach to model stock price dynamics for trading, market making, hedging, and risk management. By utilizing a time series of values, such as the price fluctuations of financial products being traded in the markets, one can construct statistical models which can assist in the extraction of useful information about the current state of the market and a set of probabilities for possible future states, such as price or volatility changes. Many models, such as the Black-Scholes-Merton model~\cite{black1973pricing}, attempted to mathematically deduce the price of options and can be used to provide useful indications of future price movements. 

However, at some point as more market participants started using the same model the behaviour of the price changed to the point that it could no longer be taken advantage of. Newer models, such as the stochastic modelling of limit order book dynamics \cite{cont2010stochastic}, the jump-diffusion processes for stock dynamics \cite{bandi2016price} and volatility estimation of market microstructure noise \cite{ait2009estimating} have been attempts predict multiple aspects of the financial markets. However such models are designed to be tractable, even at the cost of reliability and accuracy, and thus they do not necessarily fit empirical data very well.

The aforementioned properties put handcrafted models at a disadvantage, since the financial markets very frequently exhibit irrational behaviour, mainly due to the large influence of human activity, which frequently causes these models to fail. Combining Machine Learning models with handcrafted features usually improves the forecasting abilities of such models, by overcoming some of the aforementioned limitations, and improving predictions about various aspects of financial markets. This led many organizations that participate in the Financial Markets, such as Hedge Funds and investment firms, to increasingly use ML models, along with the conventional mathematical models, to make crucial decisions.

Furthermore, the introduction of electronic trading, that also led to the automation of trading operations, has magnified the volume of exchanges, producing a wealth of data. Deep Learning models are perfect candidates for analyzing such amounts of data, since they perform significantly better than the conventional Machine Learning methodologies when a large amount of data is available.  This is one of the reasons that Deep Learning is starting to have a role in analyzing the data coming from financial exchanges. \cite{kercheval2015modelling, tsantekidis2017using}

The most detailed type of data that financial exchanges are gathering is the comprehensive logs of every submitted order and event that is happening within their internal matching engine. This log can be used to reconstruct the Limit Order Book (LOB), which is explained further in Section \ref{data-section}. A basic task that can arise from this data is the prediction of future price movements of an asset by examining the current and past supply and demand of Limit Orders. This type of comprehensive logs kept by the exchanges is excessively large and traditional Machine Learning techniques, such as Support Vector Machines (SVMs) \cite{vapnik1995support}, usually cannot be applied out-of-the-box. 
Utilizing this kind of data directly with existing Deep Learning methods is also not possible due to their non-stationary nature. Prices fluctuate and suffer from stochastic drift, so in order for them to be effectively utilized by DL methods a preprocessing step is required to generate stationary features from them.

The main contribution of this work is the proposal of a set of stationary features that can be readily extracted from the Limit Order Book. The proposed features are thoroughly evaluated for predicting future mid price movements from large-scale high-frequency Limit Order data using several different Deep Learning models, ranging from simple Multilayer Perceptrons (MLPs) and CNNs to Recurrent Neural Networks (RNNs). Also we propose a novel Deep Learning model that combines the feature extraction ability of Convolutional Neural Networks (CNNs) with the Long Short Term Memory (LSTM) networks' power to analyze time series.

In Section~2 related work which employs ML models on financial data is briefly presented. Then, the dataset used is described in detail in Section~3. In Section~4 the proposed stationary feature extraction methodology is presented in detail, while in Section~5 the proposed Deep Learning methods are described. In Section~6 the experimental evaluation and comparisons are provided. Finally, conclusions are drawn and future work is discussed in Section~7.

\section{Related Work}

The task of regressing the future movements of financial assets has been the subject of many recent works such as \cite{kazem2013support, hsieh2011forecasting, lei2018wavelet}. Proven models such as GARCH are improved and augmented with machine learning component such as Artificial Neural Networks \cite{michell2018stock}. New hybrid models are employed along with Neural Networks to improve upon previous performance \cite{huang2012hybrid}. 

One of the most volatile financial markets is FOREX, the currency markets. In \cite{galeshchuk2016neural}, neural networks are used to predict the future exchange rate of major FOREX pairs such as USD/EUR. The model is tested with different prediction steps ranging from daily to yearly which reaches the conclusion that shorter term predictions tend to be more accurate. Other financial metrics, such as cash flow prediction, are very closely correlated to price prediction. 

In \cite{heaton2016deep}, the authors propose the ``Deep Portfolio Theory'' which applies autoencoders in order to produce optimal portfolios. This approach outperforms several established benchmarks, such as the Biotechnology IBB Index. Likewise in \cite{takeuchi2013applying}, another type of autoencoders, known as Restricted Boltzmann Machine (RBM), is applied to encode the end-of-month prices of stocks. Then, the model is fine-tuned to predict whether the price will move more than the median change and the direction of such movement. This strategy is able to outperform a benchmark momentum strategy in terms of annualized returns.

Another approach is to include data sources outside the financial time series, e.g., \cite{xiong2015deep}, where phrases related to finance, such as ``mortgage'' and ``bankruptcy'' were monitored on the Google trends platform and included as an input to a recurrent neural network along with the daily S\&P 500 market fund prices. The training target is the prediction of the future volatility of the market fund's price. This approach can greatly outperform many benchmark methods, such as the autoregressive GARCH and Lasso techniques.

The surge of DL methods has dramatically improved the performance over many conventional machine learning methods on tasks, such as speech recognition \cite{graves2013speech}, image captioning\cite{xu2015show, mao2014deep}, and question answering \cite{zhu2016visual7w}. The most important building blocks of DL are the Convolutional Neural Networks (CNN) \cite{lecun1995convolutional}, and the Recurrent Neural Networks (RNNs). Also worth mentioning is the improvement of RNNs with the introduction of Long Short-Term Memory Units (LSTMs) \cite{hochreiter1997long}, which has made the analysis of time series using DL easier and more performant. 

Unfortunately DL methods are prone to overfit especially in tasks such as price regression and many works exist trying to prevent such overfitting \cite{niu2012short, xi2014new}. Some might attribute overfitting to the lack of huge amounts of data that other tasks such as image and speech processing have available to them. A very rich data source for financial forecasting is the Limit Order Book. One of the few applications of ML in high frequency Limit Order Book data is \cite{kercheval2015modelling}, where several handcrafted features are created, including price deltas, bid-ask spreads and price and volume derivatives. An SVM is then trained to predict the direction of future mid price movements using all the handcrafted features. In \cite{tran2017temporal} a neural network architecture incorporating the idea of bilinear projection augmented with a temporal attention mechanism is used to predict LOB mid price.

Similarly in \cite{ntakaris2018mid, tran2017tensor} utilize the Limit Order Book data along with ML methods such as multilinear methods and smart feature selection to predict the future price movements. In our previous work~\cite{tsantekidis2017forecasting, tsantekidis2017using, passalis2017time} we introduced a large-scale high-frequency Limit Order Book dataset, that is also used in this paper, and we employed three simple DL models, the Convolutional Neural Networks (CNN), the Long-Short Term Memory Recurrent Neural Networks (LSTM RNNs) and the  Neural Bag-of-Features (N-BoF) model, to tackle the problem of forecasting the mid price movements. However, these approaches directly used the non-stationary raw Order Book data, making them vulnerable to distribution shifts and harming their ability to generalize on unseen data, as we also experimentally demonstrate in this paper.

To the best of our knowledge this is the first work that proposes a structured approach for extracting stationary price features from the Limit Order Book that can be effectively combined with Deep Learning models. We also provide an extensive evaluation of the proposed methods on a large-scale dataset with more than 4 million events. Also, a powerful model, that combines the CNN feature extraction properties with the LSTM's time series modelling capabilities, is proposed in order to improve the accuracy of predicting the price movement of stocks. The proposed combined model is also compared with the previously introduced methods using the proposed stationary price features.

\section{Limit Order Book Data}
\label{data-section}

In an order-driven financial market, a market participant can place two types of buy/sell orders. By posting a {\em limit order}, a trader promises to buy (sell) a certain amount of an asset at a specified price or less (more). The limit order book compromises on the valid limit order that are not executed or cancelled yet. 

This Limit Order Book (LOB) contains all existing buy and sell orders that have been submitted and are awaiting to be executed. A limit order is placed on the queue at a given price level, where, in the case of standard limit orders, the execution priority at a given price level is dictated by the arrival time (first in, first out). A {\em market order} is is an order to immediately buy/sell a certain quantity of the asset at the best available price in the limit order book. If the requested price of a limit order is far from the best prices, it may take a long time for the execution of the limit order, in which case, the order can finally be cancelled by the trader. The orders are split between two sides, the bid (buy) and the ask (sell) side. Each side contains the orders sorted by their price, in descending order for the bid side and ascending order for the ask side. 
\newcommand{\prt}{\rho}
\newcommand{\vola}{\upnu}

Following the notation used in \cite{cont2010stochastic}, a price grid is defined as $\{\prt^{(1)}(t),\dots,\prt^{(n)}(t)\}$, where $\prt^{(j)}(t) > \prt^{(i)}(t)$ for all $j>i$. The price grid contains all possible prices and each consecutive price level is incremented by a single tick from the previous price level. The state of the order book is a continuous-time process $v(t) \equiv \left(v^{(1)}(t), v^{(2)}(t), \dots, v^{(n)}(t) \right)_{t \geq 0}$, where $|v^{(i)}(t)|$ is the number of outstanding limit orders at price $\prt^{(i)}(t)$, $1 \leq i \leq n$. If $v^{(i)}(t) < 0$, then there are $-v^{(i)}(t)$ bid orders at price $\prt^{(i)}(t)$; if $v^{(i)}(t)>0$, then there are $v^{(i)}(t)$ ask orders at price $\prt^{(i)}(t)$. That is, $v^{(i)}(t) > 0$ refers to ask orders and $v^{(i)}(t) < 0$ bid orders.

The location of the best ask price in the price grid is defined by:
\[
i_a^{(1)}(t) = \inf\{i = 1, \dots, n\ ;\ v^{(i)}(t)>0 \},
\]
and, correspondingly, the location of the best bid price is defined by:
\[
i_b^{(1)}(t) = \sup\{i = 1, \dots, n\ ;\ v^{(i)}(t)<0 \}.
\]
For simplicity, we denote the best ask and bid prices as $p_a^{(1)}(t) \equiv \prt^{\left(i_a^{(1)}(t) \right)}(t)$ and $p_b^{(1)}(t) \equiv \prt^{\left(i_b^{(1)} (t)\right)}(t)$, respectively. Notice that if there are no ask (bid) orders in the book, the ask (bid) price is not defined. 

More generally, given that the $k$th best ask and bid prices exist, their locations are denoted as $i_a^{(k)}(t) \equiv i_a(t) + k-1$ and $i_b^{(k)}(t) \equiv i_b(t) + k-1$. The $k$th best ask and bid prices are correspondingly denoted by $p_a^{(k)}(t) \equiv \prt^{\left(i_a^{(k)}(t) \right)}(t)$ and $p_b^{(k)}(t) \equiv \prt^{\left(i_b^{(k)}(t) \right)}(t)$, respectively. Correspondingly, we denote the number of outstanding limit orders at the $k$th best ask and bid levels by $\vola_a^{(k)}(t) \equiv v^{\left(i_a^{(k)}(t)\right)}(t)$ and $\vola_b^{(k)}(t) \equiv v^{\left(i_b^{(k)}(t)\right)}(t)$, respectively.

Limit Order Book data can be used for a variety of tasks, such as the estimation of the future price trend or the regression of useful metrics, like the price volatility. Other possible tasks may include the early prediction of anomalous events, like extreme changes in price which may indicate manipulation in the markets. These examples are a few of multiple applications which can aid investors to protect their capital when unfavourable conditions exist in the markets or, in other cases, take advantage of them to profit.

Most modern methods that utilize financial time series data employ subsampling techniques, such as the well-known OHLC (Open-High-Low-Close) candles \cite{yang2000drift}, in order to reduce the number of features of each time interval. Although the OHLC candles preserve useful information, such as the market trend and movement ranges within the specified intervals, it removes possibly important microstructure information. Since the LOB is constantly receiving new orders in inconsistent intervals, it is not possible to subsample time-interval features from it in a way that preserves all the information it contains. This problem can be addressed, to some extent,  using recurrent neural network architectures, such as LSTMs, that are capable of natively handling inputs of varying size. This allows to directly utilize the data fully without using a time interval-based subsampling.

The LOB data used in this work is provided by Nasdaq Nordic and consists of 10 days worth of LOB events for 5 different Finnish company stocks, namely Kesko Oyj, Outokumpu Oyj, Sampo, Rautaruukki and Wärtsilä Oyj \cite{ntakaris2017benchmark,siikanen2016limit}. The exact time period of the gathered data begins from the 1st of June 2010 to the 14th of June 2010. Also, note that trading only happens during business days.

The data consists of consecutive snapshots of the LOB state after each state altering event takes place. This event might be an order insertion, execution or cancellation and after it interacts with the LOB and change its state a snapshot of the new state is taken. The LOB depth of the data that are used is $10$ for each side of Order Book, which ends up being 10 active orders (consisting of price and volume) for each side adding up to a total of $40$ values for each LOB snapshot. This ends up summing to a total of $4.5$ million snapshots that can be used to train and evaluate the proposed models.

In this work the task we aim to accomplish is the prediction of price movements based on current and past changes occurring in the LOB. This problem is formally defined as follows: Let $\mathbf{x}(t) \in \mathbb{R}^q$ denote the feature vector that describes the condition of the LOB at time $t$ for a specific stock, where $q$ is the dimensionality of the corresponding feature vector. The direction of the mid-price of that stock is defined as $l_k(t) = \{-1, 0, 1\}$ depending on whether the mid price decreased (-1), remained stationary (0) or increased (1) after $k$ LOB events occurred.
The number of orders $k$ is also called \textit{prediction horizon}. We aim to learn a model $f_k(\mathbf{x}(t))$, where $f_k: \mathbb{R}^{n} \rightarrow \{-1, 0, 1\} $, that predicts the direction $l_{k}(t)$ of the mid-price after $k$ orders.
In the following Section the aforementioned features and labels, as well as the procedure to calculate them are explained in depth. 

\section{Stationary Feature and Label Extraction}

The raw LOB data cannot be directly used for any ML task without some kind of preprocessing. The order volume values can be gathered for all stocks' LOBs and normalized together, since they are expected to follow the same distribution. However, this is not true for price values, since the value of a stock or asset may fluctuate and increase with time to never before seen levels. This means that the statistics of the price values can change significantly with time, rendering the price time series non-stationary.

Simply normalizing all the price values will not resolve the non-stationarity, since there will always be unseen data that may change the distribution of values to ranges that are not present in the current data. We present two solutions for this problem, one used in past work where normalization is applied constantly using past available statistics and a new approach to completely convert the price data to stationary values.

\subsection{Input Normalization}
\label{sec:input-normalization}

The most common normalization scheme is standardization (z-score):
\begin{equation}
x_{\text{norm}} = \dfrac{{x} - \bar{x}}{\sigma_{\bar{x}}}
\label{zscore-eq},
\end{equation}
where ${x}$ is a feature to be normalized, $\bar{x}$ is the mean and $\sigma_{\bar{x}}$ is the standard deviation across all samples. Such normalization is separately applied to the order size values and the price values. Using this kind of ``global'' normalization allows the preservation of the different scales between prices of different stocks, which we are trying to avoid. The solution presented in \cite{tsantekidis2017forecasting,tsantekidis2017using} is to use z-score to normalize each stock-day worth of data with the means and standard deviations calculated using previous day's data of the same stock. This way a major problem is avoided which is the distribution shift in stock prices, that can be caused by events such as stock splits or the large shifts in price that can happen over longer periods of time.

Unfortunately this presents another important issue for learning. The difference between the price values in different LOB levels are almost always minuscule. Since all the price levels are normalized using z-score with the same statistics, extracting features at that scale is hard. In this work we propose a novel approach to remedy this problem. Instead of normalizing the raw values of the LOB depth, we modify the price values to be their percentage difference to the current mid price of the Order Book. This removes the non-stationarity from the price values, makes the feature extraction process easier and significantly improves the performance of ML models, as it is also experimentally demonstrated in Section~\ref{sec:experiments}. To compensate for the removal of the price value itself we add an extra value to each LOB depth sample which is the percentage change of the mid price since the previous event. 

The mid-price is defined as the mid-point between the best bid and the best ask prices at time $t$ by
\begin{equation}
p_m^{(1)} (t) = \dfrac{p_a^{(1)}(t) + p_b^{(1)}(t)}{2} 
\label{mid-price-def}.
\end{equation}
Let 
\begin{align}
{p'}_a^{(i)}(t) =& \dfrac{p_a^{(i)}(t)}{p_m(t)} - 1, 	\label{stationary-price-a} \\
{p'}_b^{(i)}(t) =& \dfrac{p_b^{(i)}(t)}{p_m(t)} - 1,	\label{stationary-price-b}
\end{align}
and
\begin{equation}
{p'}_m(t) = \dfrac{p_m(t)}{p_m(t-1)} - 1.	\label{mid-price-change-def}
\end{equation}
Equations (\ref{stationary-price-a}) and (\ref{stationary-price-b}) serve as statistic features that represent the proportional difference between $i$th price and the mid-price at time $t$. Equation (\ref{mid-price-def}), on the other hand, serves as a dynamic feature that captures the proportional mid-price movement over the time period (that is, it represents asset's return in terms of mid-prices). 

We also use the cumulative sum of the sizes of the price levels as a feature, also know as Total Depth:
\begin{align}
\vola'^{(k)}_a(t) =& \sum_{i=1}^k{\vola_a^{(i)}(t)} 
\vspace{0.1cm} 	\label{size-cumsum-a}\\
\vola'^{(k)}_b(t) =& \sum_{i=1}^k{\vola_b^{(i)}(t)} 
\label{size-cumsum-b}
\end{align}
where $\vola^{(i)}_a(t)$ is number of outstanding limit order at the $i$th best ask price level and $\vola^{(i)}_b(t)$ is number of outstanding limit order at the $b$th best ask price level. 

The proposed stationary features are briefly summarized in Table \ref{features-table}. After constructing these three types of stationary features, each of them is separately normalized using standardization (z-score), as described in (\ref{zscore-eq}), and concatenated into a single feature vector $\myvec{x}_t$, where $t$ denotes the time step.

The input used for the time-aware models, such as the CNN, LSTM and CNN-LSTM, is the sequence of vectors $\myvec{X} = \{\myvec{x}_0, \myvec{x}_1, \dots , \myvec{x}_w\}$, where $w$ is the number of total number of events each one represented by a different time step input. For the models that need all the input into a single vector, such as the SVM and MLP models, the matrix $\myvec{X}$ is flatten into a single dimension so it can be used as input for these models.

\begin{table}[t]
	\caption{Brief description of each proposed stationary feature}
	\label{features-table}
	\begin{center}
		\begin{tabular}{ | c | c|}
			\hline
			\textbf{Feature} & \textbf{Description} \\
			\hline\hline
			Price level difference & \parbox[c]{10cm}{\vspace{0.2em}The difference of each price level to the current mid price, see Eq. (\ref{stationary-price-a}),(\ref{stationary-price-b}) 
				\[{p'}^{(i)}(t) = \dfrac{p^{(i)}(t)}{p_m(t)} - 1 \]
			} \\ 
			\hline
			Mid price change & \parbox[c]{10cm}{\vspace{0.2em} The change of the current mid price to the mid price of the previous time step, see Eq. (\ref{mid-price-change-def}) \\
				\[
				{p'}_m(t) = \dfrac{p_m(t)}{p_m(t-1)} - 1
				\]
			} \\
			\hline
			Depth size cumsum & \parbox[c]{10cm}{ \vspace{0.2em} Total depth at each price level, see Eq. (\ref{size-cumsum-a}), (\ref{size-cumsum-b}) 
				\[
				\vola'^{(k)}(t) = \sum_{i=1}^k{\vola^{(i)}(t)} 
				\]
			} \\ 
			\hline
		\end{tabular}
	\end{center}
\end{table}

\subsection{Labels}
\label{sec:labels}
The proposed models aim to predict the future movements of the mid price. Therefore, the ground truth labels must be appropriately generated to reflect the future mid price movements. Note that the mid price is a ``virtual'' value and no order can be guaranteed to immediately executed if placed at that exact price. However being able to predict its upwards or downwards movement provides a good estimate of the price of the future orders. A set of discrete choices must be constructed from our data to use as target for our classification models. The labels for describing the movement denoted by $y_t \in \{-1, 0, 1\}$, where $t$ denotes the timestep.

Simply using $p_m(t + k) > p_m(t)$ to determine the upward direction of the mid price would introduce unmanageable amount of noise, since the smallest change would be registered as an upward or downward movement. To remedy this, in our previous work \cite{tsantekidis2017forecasting, tsantekidis2017using} the noisy changes of the mid price were filtered by employing two averaging filters. One averaging filter was used on a window of size $k$ of the past values of the mid price and another averaging was applied on a future window $k$:
\begin{align}
m_b(t) =& \dfrac{1}{k+1} \sum_{i=0}^k p_m(t-i) \label{m-b} \\
m_a(t) =& \dfrac{1}{k} \sum_{i=1}^k p_m(t+i) \label{m-a}
\end{align}
where $p_t$ is the mid price as described in Equation~(\ref{mid-price-def}).
The label $l_t$, that expresses the direction of price movement at time $t$, is extracted by comparing the previously defined quantities ($m_b$ and $m_a$). However, using the $m_b$ values to create labels for the samples, as in \cite{tsantekidis2017forecasting, tsantekidis2017using}, is making the problem significantly easier and predictable due to the slower adaptation of the mean filter values to sudden changes in price. Therefore, in this work we remedy this issue by replacing $m_b$ with the mid price. Therefore, the labels are redefined as:
\begin{equation}
l_t =
\begin{cases}
\ \ 1, & \text{if } \dfrac{m_a(t)}{p_m(t)} > 1 + \alpha
\vspace{0.2cm}\\
-1, & \text{if } \dfrac{m_a(t)}{p_m(t)} < 1 - \alpha
\vspace{0.2cm}\\
\ \ 0, & \text{otherwise}
\end{cases}
\label{direction-eq}
\end{equation}

where $\alpha$ is the threshold that determines how significant a mid price change $m_a(t)$ must be in order to label the movement as upward or downward. Values that do not satisfy this inequality are considered as insignificant and are labeled as having no price movement, or in other words being ``stationary''. The resulting labels present the trend to be predicted. This process is applied across all time steps of the dataset to produce labels for all the depth samples.

\section{Machine Learning Models}
In this section we explain the particular inner workings of the CNN and LSTM models that are used and present how they are combined to form the proposed CNN-LSTM model. The technical details of each model are explained along with the employed optimization procedure.

\begin{figure}
	\centering
	\includegraphics[scale=0.4]{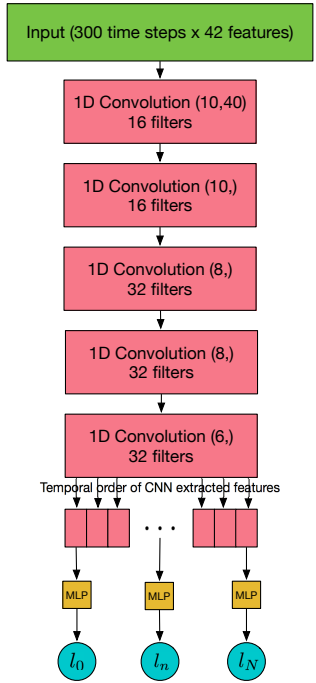}
	\caption{A visual representation of the evaluated CNN model. Each layer includes the filter input size and the number of filters used.}
	\label{fig:cnn-model}
\end{figure}

\subsection{Convolutional Neural Networks}
\label{sec:conv-nets}

Convolutional Neural Networks (CNNs) consist of the sequential application of convolutional and pooling layers usually followed by some fully connected layers, as shown in Figure~\ref{fig:cnn-model}. Each convolutional layer $i$ is equipped with a set of filters $\mathbf{W}_i \in \mathbb{R} ^{S \times D \times N}$ that is convolved with an input tensor, where $S$ is the number of used filters, $D$ is the {filter size},  and $N$ is the number of the input channels. The input tensor $\mathbf{X} \in \mathbb{R}^{(B \times T \times F)}$ is consisted by the temporally ordered features described in Section \ref{sec:input-normalization}, where $B$ is the batch size, $T$ is the number of time steps and $F$ is the number of features per time step.

In this work we leverage the causal padding introduced in \cite{van2016wavenet} to avoid using future information to produce features for the current time step. Using a series of convolutional layers allows for capturing the fine temporal dynamics of the time series as well as correlating temporally distant features. After the last convolutional/pooling layer a set of fully connected layers are used to classify the input time series. The network's output expresses the categorical distribution for the three direction labels (upward, downward and stationary), as described in (\ref{direction-eq}), for each time-step.

We also employ a temporal batching technique, similar to the one used in LSTMs, to increase the computational efficiency and reduce memory requirements of our experiments when training with CNNs. Given the above described input tensor $\myvec{X}$ and convolution filters $\myvec{W}_i$ the last convolution produces a tensor with dimensions $ (B,T,S,N) $, which in most uses cases is flattened to a tensor of size $(B, T \times S \times N)$ before being fed to a fully connected layer. Instead we retain the temporal ordering by only reducing the tensor to dimension $(B, T, S \times N) $. An identical fully connected network with a softmax output is applied for each $S \times N$ vectors leading to $T$ different predictions. 

Since we are using causal convolutions with "full" padding, all the convolutional layers produce the same time steps $T$, hence we do not need to worry about label alignment to the correct time step. Also the causal convolutions ensure that no information from the future leaks to past time step filters. This technique reduces the receptive field of the employed CNN, but this can be easily remedied by using a greater number of convolutional layers and/or a larger filter size $D$.

\subsection{Long Short Term Memory Recurrent Neural Networks}

One of the most appropriate Neural Network architectures to apply on time series is the Recurrent Neural Network (RNN) architecture. Although powerful in theory, this type of network suffers from the vanishing gradient problem, which makes the gradient propagation through a large number of steps impossible. An architecture that was introduced to solve this problem is the Long Short Term Memory (LSTM) networks~\cite{hochreiter1997long}. This architecture protects its hidden activation from the decay of unrelated inputs and gradients by using gated functions between its ``transaction'' points. The protected hidden activation is the ``cell state''  which is regulated by said gates in the following manner:

\begin{align}
\myvec{f}_t &= \sigma(\myvec{W}_{xf} \cdot \myvec{x} + \myvec{W}_{hf} \cdot \myvec{h}_{t-1} + \myvec{b}_f) \\
\myvec{i}_t &= \sigma(\myvec{W}_{xi} \cdot \myvec{x} + \myvec{W}_{hi} \cdot \myvec{h}_{t-1} + \myvec{b}_i) \\
\myvec{c}'_t &= tanh(\myvec{W}_{hc} \cdot \myvec{h}_{t-1} + \myvec{W}_{xc} \cdot \myvec{x}_t + \myvec{b}_c) \\
\myvec{c}_t &= \myvec{f}_t \cdot \myvec{c}_{t-1} + \myvec{i}_t \cdot \myvec{c}'_t  \\
\myvec{o}_t &= \sigma(\myvec{W}_{oc} \cdot \myvec{c}_t + \myvec{W}_{oh} \cdot \myvec{h}_{t-1} + \myvec{b}_o) \\
\myvec{h}_t &= \myvec{o}_t \cdot \sigma(\myvec{c}_t) 
\end{align}
where $\myvec{f}_t$, $\myvec{i}_t$ and $\myvec{o}_t$ are the activations of the input, forget and output gates at time-step $t$, which control how much of the input and the previous state will be considered and how much of the cell state will be included in the hidden activation of the network. The protected cell activation at time-step $t$ is denoted by $\myvec{c}_t$, whereas $\myvec{h}_t$ is the activation that will be given to other components of the model. The matrices $\myvec{W}_{xf}, \myvec{W}_{hf}, \myvec{W}_{xi}, \myvec{W}_{hi}, \myvec{W}_{hc}, \myvec{W}_{xc}, \myvec{W}_{oc}, \myvec{W}_{oh}$ are used to denote the weights connecting each of the activations with the current time step inputs and the previous time step activations.

\subsection{Combination of models (CNN-LSTM)}

We also introduce a powerful combination of the two previously described models. The CNN model is identically applied as described in Section \ref{sec:conv-nets}, using causal convolutions and temporal batching to produce a set of features for each time step. In essence the CNN acts as the feature extractor of the LOB depth time series, which produces a new time series of features with the same length as the original one, with each of them having time steps corresponding to one another.

\begin{figure}
	\centering
	\includegraphics[scale=0.3]{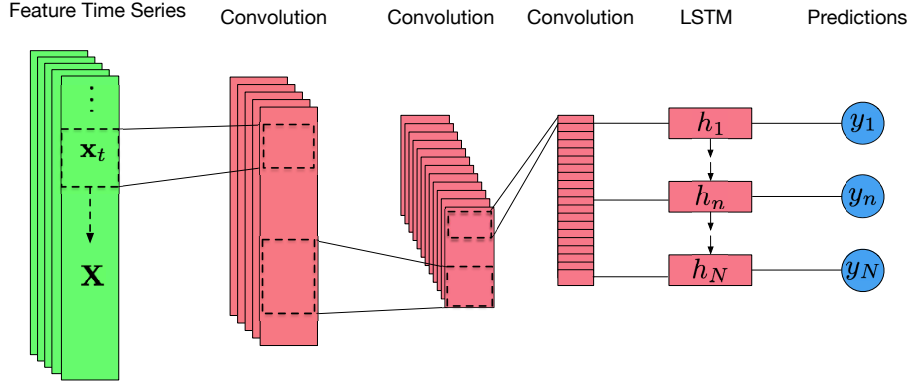}
	\caption{CNN-LSTM model}
	\label{fig:cnnlstm}
\end{figure}

An LSTM layer is then applied on the time series produced by the CNN, and in turn produces a label for each time step. This works in a very similar way to the fully connected layer described in \ref{sec:conv-nets} for temporal batching, but instead of the Fully Connected layer the LSTM allows the model to incorporate the features from past steps. The model architecture is visualized in Figure~\ref{fig:cnnlstm}.

\subsection{Optimization}
\label{sec:optimization}
The parameters of the models are learned by minimizing the categorical cross entropy loss defined as: 
\begin{equation}
\mathcal{L}(\myvec{W}) = -\sum_{i=1}^{L} y_i \cdot \log \hat{y}_i,
\end{equation}
where $L$ is the number of different labels and the notation $\myvec{W}$ is used to refer to the parameters of the models. The ground truth vector is denoted by  $\mathbf{y}$, while $\hat{\mathbf{y}}$ is the predicted label distribution. The loss is summed over all samples in each batch. Due to the unavoidable class imbalance of this type of dataset, a weighted loss is employed to improve the mean recall and precision across all classes:
\begin{equation}
\label{eq:loss}
\mathcal{L}(\myvec{W}) = -\sum_{i=1}^{L} c_{y_i} \cdot y_i \cdot \log \hat{y}_i,
\end{equation}
where $c_{y_i}$ is the assigned weight for the class of $y_i$. The individual weight   $c_i$ assigned to each class $i$ is calculated as:
\begin{equation}
c_i = \dfrac{|\mathcal{D}|}{n \cdot |\mathcal{D}_i|},
\end{equation}
where $ |\mathcal{D}| $ is the total number of samples in our dataset $\mathcal{D}$, $n$ is the total number of classes (which in our case is 3) and $\mathcal{D}_i$ is set of samples from our dataset that have been labeled to belong in class $i$.

The most commonly used method to minimize the loss function defined in (\ref{eq:loss}) and learn the parameters  $\myvec{W}$ of the model is gradient descent \cite{werbos1990backpropagation}:
\begin{equation}
\myvec{W}' = \myvec{W} - \eta \cdot \dfrac{\partial \mathcal{L}}{\partial \myvec{W}}
\end{equation}
where $\myvec{W}'$ are the parameters of the model after each gradient descent step and $\eta$ is the learning rate. In this work we utilize the RMSProp optimizer \cite{tieleman2012lecture}, which is an adaptive learning rate method and has been shown to improve the training time and performance of DL models.

The LSTM, CNN and CNN-LSTM models along with all the training algorithms were developed using Keras \cite{chollet2015keras}, which is a framework built on top of the Tensorflow library \cite{tensorflow2015-whitepaper}.

\section{Experimental Evaluation}
\label{sec:experiments}

All the models were tested for step sizes $k = 10, 50, 100,$ and $200$ in (\ref{m-a}), where the $\alpha$ value for each was set at $2 \times 10^{-5},\ 9 \times 10^{-5},\ 3 \times 10^{-4}$ and $ 3.5 \times 10^{-4} $ respectively. The parameter $\alpha$ was chosen in conjunction with the future horizon with the aim to have relatively balanced distribution of labels across classes. In a real trading scenario it is not possible to have a profitable strategy that creates as many trade signals as ``no-trade'' signals, because it would accumulate enormous commission costs. For that reason $\alpha$ is selected with the aim to get a logical ratio of about 20\% long, 20\% short and 60\% stationary labels. The effect of varying the parameter $\alpha$ on the class distribution of labels is shown in Table \ref{alpha-table}. Note that increasing the $\alpha$ allows for reducing the number of trade signals which should be changed depending on the actual commission and slippage costs that are expected to occur.

\begin{table}
	\caption{Example of sample distribution across classes depending on $\alpha$ for prediction horizon $k =100$}
	\label{alpha-table}
	\begin{center}
		\begin{tabular}{ | c |c| c| c|}
			\hline
			\hspace{2em}$\alpha$\hspace{2em} & \hspace{1em}Down\hspace{1em} & Stationary & \hspace{1.5em}Up\hspace{1.5em} \\
			\hline\hline
			$1.0 \times 10^{-5}$ & $0.39$&$0.17$&$0.45$ \\ \hline
			$2.0 \times 10^{-5}$ & $0.38$&$0.19$&$0.43$ \\ \hline
			$5.0 \times 10^{-5}$ & $0.35$&$0.25$&$0.41$ \\ \hline
			$1.0 \times 10^{-4}$ & $0.30$&$0.33$&$0.36$ \\ \hline
			$2.0 \times 10^{-4}$ & $0.23$&$0.49$&$0.28$ \\ \hline
			$3.0 \times 10^{-4}$ & $0.18$&$0.60$&$0.22$ \\ \hline
			$3.5 \times 10^{-4}$ & $0.15$&$0.66$&$0.19$ \\ \hline
		\end{tabular}
	\end{center}
\end{table}

We tested the CNN and LSTM models using the raw features and the proposed stationary features separately and compared the results. The architecture of the three models that were tested is described bellow.

The proposed CNN model consists of the following sequential layers:

\begin{center}
	\begin{minipage}{0.6\textwidth}
		\begin{enumerate}
			\item 1D Convolution with 16 filters of size $(10,42)$
			\item 1D Convolution with 16 filters of size $(10,)$
			\item 1D Convolution with 32 filters of size $(8,)$
			\item 1D Convolution with 32 filters of size $(6,)$
			\item 1D Convolution with 32 filters of size $(4,)$
			\item Fully connected layer with 32 neurons
			\item Fully connected layer with 3 neurons 
		\end{enumerate}
	\end{minipage}
\end{center}
The activation function used for all the convolutional and fully connected layer of the CNN is the Parametric Rectifying Linear Unit (PRELU) \cite{he2015delving}. The last layer uses the softmax function for the prediction of the probability distribution between the different classes. All the convolutional layers are followed by a Batch Normalization (BN) layer after them.

The LSTM network uses 32 hidden neurons followed by a feed-forward layer with 64 neurons using Dropout and PRELU as activation function. Experimentally we found out that the hidden layer of the LSTM should contain 64 or less hidden neurons to avoid over-fitting the model. Experimenting with higher number of hidden neurons would be feasible if the dataset was even larger. 

Finally the CNN-LSTM model applies the convolutional feature extraction layers on the input and then feeds them in the correct temporal order to an LSTM model. The CNN component is comprised of the following layers:
\begin{center}
	\begin{minipage}{0.6\textwidth}
		\begin{enumerate}
			\item 1D Convolution with 16 filters of size $(5,42)$ 
			\item 1D Convolution with 16 filters of size $(5,)$ 
			\item 1D Convolution with 32 filters of size $(5,)$ 
			\item 1D Convolution with 32 filters of size $(5,)$ 
		\end{enumerate}
	\end{minipage}
\end{center}

Note that the receptive field of each convolutional filter in the CNN module is smaller that the standalone CNN, since the LSTM can capture most of the information from past time steps. The LSTM module has the same exact architecture as the standalone LSTM. A visual representation of this CNN-LSTM model is shown in Figure~\ref{fig:cnnlstm}. Likewise, PRELU is the activation function used for the CNN and the fully connected layers, while the softmax function is used for the output layer of the network to predict the probability distribution of the classes.

\begin{figure}
	\centering
	\includegraphics[scale=0.70]{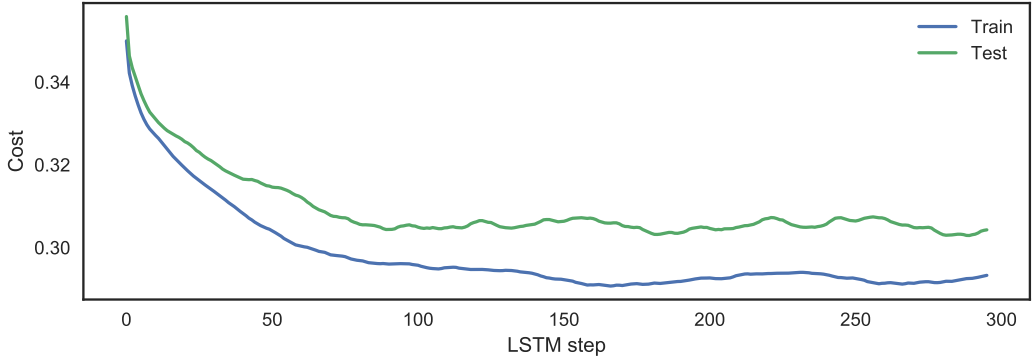}
	\caption{Mean cost per recurrent step of the LSTM network}
	\label{bad-step-score}
\end{figure}

\begin{table*}
	\caption{Experimental results for different prediction horizons $k$. The values that are reported are the mean of each metric for the last 20 training epochs.}
	\label{results-table}

	\begin{center}
\footnotesize
\bgroup
\def\arraystretch{0.8}%
		\begin{tabular}{ |c|c|c|c|c|c|c|c|c|}
			\hline

			\multirow{1}{*}{\textbf{Feature Type}} & 
			\multicolumn{1}{c|}{\textbf{Model}} &
			\multicolumn{1}{c|}{\textbf{Mean Recall}} &
			\multicolumn{1}{c|}{\textbf{Mean Precision}} &
			\multicolumn{1}{c|}{\textbf{Mean F1}} &  \multicolumn{1}{c|}{\textbf{Cohen's} $\kappa$} \\  \cline{1-6}            
			
			\multicolumn{6}{|c|}{\multirow{2}{*}{Prediction Horizon $k=10$}} \\ 
			\multicolumn{6}{|c|}{} \\ \cline{1-6}
			\multirow{4}{*}{\textbf{Raw Values}} 
			& SVM &  $0.35  $ & $0.43  $ & $0.33  $ & $0.04   $ \\ \cline{2-6}
			& MLP &  ${ 0.34   }$ & ${ 0.34   }$ & ${ 0.09   }$ & ${ 0.00   }$ \\ \cline{2-6}
			& CNN &  ${ 0.51   }$ & ${ 0.42   }$ & ${ 0.38   }$ & ${ 0.14   }$ \\ \cline{2-6}
			& LSTM &  ${ 0.49   }$ & ${ 0.41   }$ & ${ 0.35   }$ & ${ 0.12   }$ \\ \cline{1-6}

			\multirow{5}{*}{\textbf{Stationary Features}} 
			& SVM &  $0.33  $ & $\mathbf{0.46  }$ & $0.30  $ & $0.011  $ \\ \cline{2-6}
			& MLP  &  ${ 0.34   }$ & ${ 0.35   }$ & ${ 0.09   }$ & ${ 0.00   }$ \\ \cline{2-6}
			& CNN  &  ${ 0.54   }$ & ${ 0.44   }$ & ${ 0.43   }$ & ${ 0.19   }$ \\ \cline{2-6}
			& LSTM  &  ${ 0.55   }$ & ${ 0.45   }$ & ${ 0.42   }$ & ${ 0.18   }$ \\ \cline{2-6}
			& CNNLSTM  &  $\mathbf{ 0.56   }$ & ${ 0.45   }$ & $\mathbf{ 0.44   }$ & $\mathbf{ 0.21   }$ \\ \cline{1-6}
			
			\multicolumn{6}{|c|}{\multirow{2}{*}{Prediction Horizon $k=50$}} \\ 
			\multicolumn{6}{|c|}{} \\ \cline{1-6}
			
			\multirow{4}{*}{\textbf{Raw Values}} 
			& SVM &  $0.35  $ & $0.41  $ & $0.32  $ & $0.03   $ \\ \cline{2-6}
			& MLP  &  ${ 0.41   }$ & ${ 0.38   }$ & ${ 0.21   }$ & ${ 0.04   }$ \\ \cline{2-6}
			& CNN  &  ${ 0.50   }$ & ${ 0.42   }$ & ${ 0.37   }$ & ${ 0.13   }$ \\ \cline{2-6}
			& LSTM  &  ${ 0.46   }$ & ${ 0.40   }$ & ${ 0.34   }$ & ${ 0.10   }$ \\ \cline{1-6}

			\multirow{5}{*}{\textbf{Stationary Features}}
			& SVM &  $0.39   $ & $0.41   $ & $0.38   $ & $0.09    $ \\ \cline{2-6}
			& MLP &  $0.49  $ & $0.43  $ & $0.38  $ & $0.14   $ \\ \cline{2-6}
			& CNN  &  $0.55   $ & $0.45   $ & $0.43   $ & $0.20   $ \\ \cline{2-6}
			&LSTM  &  $\mathbf{0.56  } $ & $0.46  $ & $0.44   $ & $0.21  $ \\ \cline{2-6}
			& CNNLSTM &  $\mathbf{0.56  }$ & $\mathbf{0.47  }$ & $\mathbf{0.47  }$ & $\mathbf{0.24  } $ \\ \cline{1-6}

			\multicolumn{6}{|c|}{\multirow{2}{*}{Prediction Horizon $k=100$}} \\ 
			\multicolumn{6}{|c|}{} \\ \cline{1-6}
			
			\multirow{4}{*}{\textbf{Raw Values}} 
			& SVM &  $0.35  $ & $0.46  $ & $0.33  $ & $0.05   $ \\ \cline{2-6}
			& MLP  &  ${ 0.45   }$ & ${ 0.39   }$ & ${ 0.26   }$ & ${ 0.06   }$ \\ \cline{2-6}
			& CNN  &  ${ 0.49   }$ & ${ 0.42   }$ & ${ 0.37   }$ & ${ 0.12   }$ \\ \cline{2-6}
			& LSTM  &  ${ 0.45   }$ & ${ 0.39   }$ & ${ 0.34   }$ & ${ 0.09   }$ \\ \cline{1-6}
			
			\multirow{5}{*}{\textbf{Stationary Features}}
			& SVM &  $0.36  $ & $0.46  $ & $0.35  $ & $0.07   $ \\ \cline{2-6}
			& MLP  &  ${ 0.50   }$ & ${ 0.43   }$ & ${ 0.39   }$ & ${ 0.14   }$ \\ \cline{2-6}
			& CNN  &  ${ 0.54   }$ & ${ 0.46   }$ & ${ 0.44   }$ & ${ 0.21   }$ \\ \cline{2-6}
			& LSTM  &  $\mathbf{ 0.56   }$ & ${ 0.46   }$ & ${ 0.44   }$ & ${ 0.20   }$ \\ \cline{2-6}
			& CNNLSTM  &  ${ 0.55   }$ & $\mathbf{ 0.47   }$ & $\mathbf{ 0.48   }$ & $\mathbf{ 0.24   }$ \\ \cline{1-6}

			\multicolumn{6}{|c|}{\multirow{2}{*}{Prediction Horizon $k=200$}} \\ 
			\multicolumn{6}{|c|}{} \\ \cline{1-6}
			
			\multirow{4}{*}{\textbf{Raw Values}} 
			& SVM &  $0.35  $ & $0.44  $ & $0.31  $ & $0.04   $ \\ \cline{2-6}
			& MLP  &  ${ 0.44   }$ & ${ 0.40   }$ & ${ 0.32   }$ & ${ 0.08   }$ \\ \cline{2-6}
			& CNN  &  ${ 0.47   }$ & ${ 0.43   }$ & ${ 0.39   }$ & ${ 0.14   }$ \\ \cline{2-6}
			& LSTM  &  ${ 0.42   }$ & ${ 0.39   }$ & ${ 0.36   }$ & ${ 0.08   }$ \\ \cline{1-6}

			\multirow{5}{*}{\textbf{Stationary Features}}
			& SVM &  $0.38   $ & $0.46   $ & $0.36   $ & $0.10   $ \\ \cline{2-6}
			& MLP  &  ${ 0.49   }$ & ${ 0.45   }$ & ${ 0.42   }$ & ${ 0.17   }$ \\ \cline{2-6}
			& CNN  &  ${ 0.51   }$ & ${ 0.47   }$ & ${ 0.45   }$ & ${ 0.20   }$ \\ \cline{2-6}
			& LSTM  &  ${ 0.52   }$ & ${ 0.47   }$ & ${ 0.46   }$ & ${ 0.22   }$ \\ \cline{2-6}
			& CNNLSTM  &  $\mathbf{ 0.53   }$ & $\mathbf{ 0.48   }$ & $\mathbf{ 0.49   }$ & $\mathbf{ 0.25   }$ \\ \cline{1-6}

		\end{tabular}
\egroup
	
	\end{center}
\end{table*}

One recurring effect we observe when training LSTM networks on LOB data is that for the first steps of observation the predictions $y_i$ yield a bigger cross entropy cost, meaning worse performance in our metrics. We run a set of experiments where the LSTM was trained for all the steps of the input windows $T$. The resulting mean cost per time step can be observed in Figure \ref{bad-step-score}. As a result, trying to predict the price movement using insufficient past information is not possible and should be avoided since it leads to noisy gradients. To avoid this, a ``burn-in'' input is initially used to build its initial perception of the market before actually making correct decisions. In essence the first ``burn-in'' steps of the input are skipped, by not allowing any gradient to alter our model until after the 100th time step. We also apply the same method to the CNN-LSTM model.

\begin{figure*}
	\centering
	\includegraphics[width=1.02\linewidth]{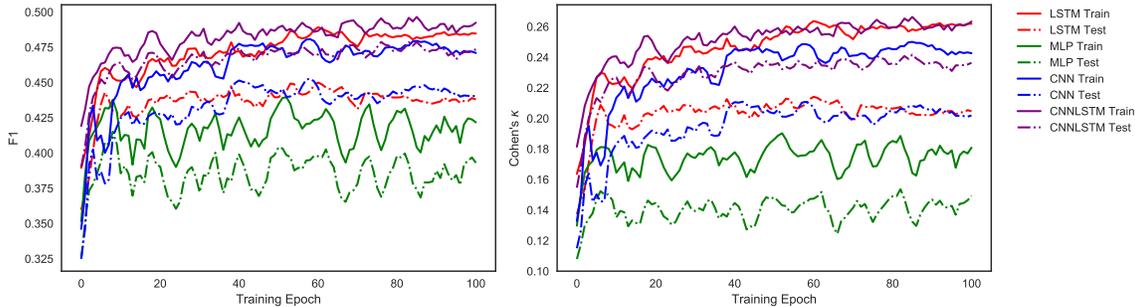}
	\caption{F1 and Cohen's $\kappa$ metrics during training for prediction horizon $k=100$. Plots are smoothed with a mean filter with window=3 to reduce fluctuations.
	}
	\label{fig:f1-kappa-training}
\end{figure*}

For training the models, the dataset is split as follows. The first 7 days of each stock are used to train the models, while the final 3 days are used as test data. The experiments were conducted for 4 different prediction horizons $k$, as defined in  (\ref{m-a}) and (\ref{direction-eq}).

Performance is measured using Cohen's kappa \cite{cohen1960coefficient}, which is used to evaluate the concordance between sets of given answers, taking into consideration the possibility of random agreements happening. The mean recall, mean precision and mean F1 score between all 3 classes is also reported. Recall is the number of true positives samples divided by the sum of true positives and false negatives, while precision is the number of true positive divided by the sum of true positives and false positives. F1 score is the harmonic mean of the precision and recall metrics.

The results of the experiments are shown in Table \ref{results-table}. The results are compared for the models trained on the raw price features with the ones trained using the extracted stationary features. The results confirm that extracting stationary features from the data significantly improve performance of Deep Learning models such as CNNs and LSTMs.

We also trained a Linear SVM model and a simple MLP model and compared them to the DL models. The SVM model was trained using Stochastic Gradient Descent since the size of the dataset is too large to use a regular Quadratic Programming solver. The SVM model implementation is provided by the sklearn library \cite{pedregosa2011scikit}. The MLP model consists of three fully connected layers with sizes 128, 64, 32, and PRELU as activations for each layers. Dropout is also used to avoid overfitting and the softmax activation function was used in the last layer.

Since both the SVM and the MLP models cannot iterate over timesteps to gain the same amount of information as the CNN and LSTM-based models, a window of 50 depth events is used and is flattened into a single sample. This process is applied in a rolling fashion for all the dataset to generate a dataset upon which the two models can be trained. One important note is the training fluctuations that are observed in Figure \ref{fig:f1-kappa-training}, which are caused by the great class imbalance. Similar issues where observed in initial experiments with CNN and LSTM models but using the weighted loss described in \ref{sec:optimization} the fluctuations subsided.

The proposed stationary price features significantly outperform the raw price features for all the tested models. This can be attributed to a great extent to the stationary nature of the proposed features. The employed price differences provide an intrinsically stationary and normalized price measure that can be directly used. This is in contrast with the raw price values that requires careful normalization to ensure that their values remain into a reasonable range and suffer for significantly non-stationarity issues when the price increases to levels not seen before. By converting the actual prices to the their difference to the mid price and normalize that, this important feature is exaggerated to avoid being suppressed by the much larger price movements through time. The proposed combination model CNN-LSTM also outperforms its separated individual component models as shown in Figure \ref{fig:f1-kappa-training} and Table \ref{results-table} showing that it can better handle the LOB data and use them to take advantage of the microstructure existing within the data to produce more accurate predictions. 

\section{Conclusion}

In this paper we proposed a novel method for extracting stationary features from raw LOB data, suitable for use with different DL models. Using different ML models, i.e., SVMs, MLPs, CNNs and LSTMs, it was experimentally demonstrated that the proposed features significantly outperform the raw price features. The proposed stationary features achieve this by making the difference between the prices in the LOB depth the main metric instead of the price itself, which usually fluctuates much more through time than the price level within the LOB. A novel combined CNN-LSTM model was also proposed for time series predictions and it was demonstrated that exhibits more stable behaviour and leads to better results that the CNN and LSTM models.

There are several interesting future research directions. As with all the DL application, more data would enable the use of bigger models that would not be at risk of being overtrained as it was observed in this work. An RNN-type of network could be also used to perform a form of ``intelligent'' r-sampling extracting useful features from a specific and limited time-interval of depth events, which would avoid losing information and allow for the later models produce prediction for a certain time period and not for a number of following events. Another important addition would be an attention mechanism \cite{xu2015show}, \cite{cho2015describing}, which would allow for the better observation of the features by the network allowing it to ignore noisy parts of the data and use only the relevant information.

\section*{Acknowledgment}
The research leading to these results has received funding from the H2020 Project BigDataFinance MSCA-ITN-ETN 675044 (http://bigdatafinance.eu), Training for Big Data in Financial Research and Risk Management.

\bibliographystyle{elsarticle-num}


\end{document}